\def\year{2020}\relax
\documentclass[letterpaper]{article} 
\usepackage{aaai20}  
\usepackage{times}  
\usepackage{helvet} 
\usepackage{courier}  
\usepackage[hyphens]{url}  
\usepackage{graphicx} 
\usepackage{graphicx}  
\usepackage{amsmath}
\usepackage{times}
\usepackage{graphicx}

\usepackage{amssymb}
\usepackage{algorithm}
\usepackage{algorithmicx}
\usepackage{algpseudocode}
\usepackage{mathrsfs}
\usepackage{multirow}
\usepackage{amsthm}
\usepackage{bm}
\usepackage{mathrsfs}

\newtheorem{theorem}{Theorem}
\newtheorem{lemma}{Lemma}
\newtheorem{assumption}{Assumption}
\newtheorem{example}{Example}
\newtheorem{remark}{Remark}

\newtheorem{definition}{Definition}
\newtheorem{corollary}{Corollary}

\usepackage{scalerel}
\usepackage{comment}

\title{Polynomial Matrix Completion for\\ Missing Data Imputation and Transductive Learning} 
\author{Jicong Fan, Yuqian Zhang, Madeleine Udell\\
Cornell University, Ithaca, NY 14853, USA\\
\{jf577, yz2557, udell\}@cornell.edu
}

\begin{document}

\maketitle

\begin{abstract}
This paper develops new methods to recover the missing entries of
a high-rank or even full-rank matrix
when the intrinsic dimension of the data is low compared to the ambient dimension.
Specifically, we assume that the columns of a matrix are generated
by polynomials acting on a low-dimensional intrinsic variable,
and wish to recover the missing entries under this assumption.
We show that we can identify the complete matrix of minimum intrinsic dimension
by minimizing the rank of the matrix in a high dimensional feature space.
We develop a new formulation of the resulting problem
using the kernel trick together with a new relaxation of the rank objective,
and propose an efficient optimization method.
We also show how to use our methods to complete data drawn from multiple nonlinear manifolds.
Comparative studies on synthetic data, subspace clustering with missing data,
motion capture data recovery, and transductive learning verify the superiority of our methods over the state-of-the-art.
\end{abstract}

\section{Introduction}
The low-rank matrix completion (LRMC) problem is to recover the missing entries of a partially observed matrix of low-rank  \cite{CandesRecht2009}. 
Suppose matrix $\bm{X}\in\mathbb{R}^{m\times n}$ is low-rank: $\textup{rank}(\bm{X})=r\ll \min\lbrace m,n\rbrace$. The observed entries of $\bm{X}$ are denoted by $\lbrace X_{ij}\rbrace_{(i,j)\in\Omega}$, where $\Omega$ consists of the locations of observed entries. Define $\mathcal{P}_{\Omega}(X_{ij})=X_{ij}$ if $(i,j)\in\Omega$ and $\mathcal{P}_{\Omega}(X_{ij})=0$ if $(i,j)\notin\Omega$. A typical LRMC method is 
\begin{equation}\label{Eq.LRMC_nnm}
\mathop{\text{minimize}}_{\hat{\bm{X}}}\Vert\hat{\bm{X}}\Vert_\ast,\ \textup{subject to}\ \mathcal{P}_{\Omega}(\hat{\bm{X}})=\mathcal{P}_{\Omega}(\bm{X}),
\end{equation}
where $\Vert\hat{\bm{X}}\Vert_\ast$ denotes the nuclear norm of $\hat{\bm{X}}$ . The nuclear norm is a convex relaxation of rank function (NP-hard) and defined by the sum of singular values of matrix.
\cite{CandesRecht2009} proved that if the number of observed entries $\vert\Omega\vert$ obeys $\vert\Omega\vert\geq Cn^{1.2}r\log n$ for some positive constant $C$, $\bm{X}$ can be exactly recovered with high probability via solving \eqref{Eq.LRMC_nnm}. 
Many other LRMC methods based on low-rank factorization or nonconvex regularizations can be found in \cite{MC_MF_Wen2012,MC_TNNR_PAMI2013,MC_ShattenP_AAAI125165,gu2014weighted,MC_IRRN,icml2014c2_wanga14,mu2016scalable,xie2016weighted,sun2016guaranteed}. The applications of LRMC include recommendation system \cite{Su:2009:SCF:1592474.1722966}, image inpainting \cite{imageinpainting2014}, classification \cite{NIPS2010_3932}, and so on.

\begin{figure}[ht]
\includegraphics[width=7cm]{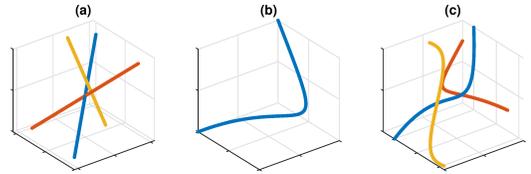}
\centering
\caption{Examples of data forming high-rank matrices in 3D space: (a) multiple subspaces; (b) one nonlinear manifold; (c) multiple nonlinear manifolds.}
\label{fig.MC4}
\end{figure}

The low-rank assumption \cite{udell2019big} implies that LRMC methods cannot effectively handle high-rank matrices (those cannot be well approximated by low-rank matrices) even if the intrinsic dimensions of the data are low. High-rank matrices with low intrinsic dimension are pervasive and often resulted from multiple-subspace sampling or nonlinear mappings. Figure \ref{fig.MC4} shows three examples. The models have been proposed for subspace clustering \cite{SSC_PAMIN_2013}, manifold learning \cite{roweis2000nonlinear,van2009dimensionality}, and deep learning \cite{hinton2006reducing}.

\textbf{Related work} Recently, matrix completion on multiple-subspace data and nonlinear data has drawn many researchers' attention \cite{ErikssonBalzanoNowak2011,YangRobinsonVidal,li2016structured,fan2018DMF,NLMC2016,NIPS2016_6357,fan2018matrix,Fan2017290,pmlr-v70-ongie17a,FANNLMC,LADMC}. For example, \cite{NIPS2016_6357} proposed a group-sparse optimization with rank-one constraints to complete and cluster multiple-subspace data. \cite{pmlr-v70-ongie17a} and \cite{FANNLMC}  proposed to perform rank minimization in the high-dimensional feature space induced by kernels to recover the missing entries of data generated by nonlinear models.
Although these matrix completion methods can outperform
low-rank matrix completion methods, 
the theory is lacking:
for example, existing sampling complexity bounds are not rigorous.
Moreover, our numerical results indicate that
there is significant room to improve recovery accuracy.

\textbf{Contributions} In this paper, we propose 1) a new generative model for high-rank matrices: finite degree polynomials map a low-dimensional latent variable to a high-dimensional ambient space; 2) two high-rank matrix completion methods; and 3) an efficient algorithm to solve the optimization. Furthermore, we analyze the sample complexity of the high-rank model, which further confirms the superiority of the proposed methods over LRMC methods. We applied the proposed methods to subspace clustering on incomplete data, motion capture data recovery, and transductive learning, and achieved state-of-the-art results.

\textbf{Notation}
\begin{table}[h!]
\centering
\begin{tabular}{ll}
\hline
$\sigma_i(\bm{Y})$ &the $i$-th singular value of matrix $\bm{Y}$ \\
$\textup{Tr}(\bm{Y})$ & the trace of $\bm{Y}$\\
$\bm{y}_j$ &the $j$-th column of $\bm{Y}$ \\
$\Vert\bm{Y}\Vert_\infty$ &$\max_{ij}\vert Y_{ij}\vert$\\
$f^{(k)}$& the $k$-th order derivative of function $f$\\
$\bm{I}_s$& $s\times s$ identity matrix\\
$\odot$	&Hadamard product \\
\hline
\end{tabular}
\end{table}

\section{Polynomial Matrix Completion (PMC)}
In this paper, we assume the data matrix $\bm{X} \in \mathbb{R}^{d\times n}$
is generated by the following model.
\begin{assumption}\label{asp_fz_analytic}
Suppose $d\ll m\ll n$, $\bm{Z}\in\mathbb{R}^{d\times n}$ is full-rank, $\Vert \bm{Z}\Vert_\infty\leq c_z$, and $f:\mathbb{R}^d\rightarrow\mathbb{R}^m$ is analytic. For $j=1,2,\ldots,n$, let $\bm{x}_j=f(\bm{z}_j)$.
\end{assumption}
The matrix $\bm{X}$ generated by Assumption \ref{asp_fz_analytic}
can be of high-rank or even full-rank if $f$ is nonlinear,
even though the intrinsic dimension $d$ of the data is much lower than the ambient dimension $m$.
In order to exploit the low intrinsic dimension,
we will make use of a polynomial feature map:
\begin{definition}\label{def_phi}
Let $\lbrace c_\mu\rbrace_{\vert\mu\vert\leq q}$ be nonzero parameters. A $q$-order polynomial feature map of $\bm{x}\in\mathbb{R}^m$ is defined as
$$\phi(\bm{x})=(c_{\mu}x_1^{\mu_1}x_2^{\mu_2}\cdots x_m^{\mu_m})_{\vert\mu\vert\leq q}\in\mathbb{R}^l,$$
where $\lbrace\mu_i\rbrace_{i=1}^m$ are non-negative integers, $\vert\mu\vert=\mu_1+\mu_2+\cdots\mu_m$, and $l=\binom{m+q}{q}$.
\end{definition}

Let $\phi$ be a $q$-order polynomial feature map and denote $\phi(\bm{X})=[\phi(\bm{x}_1),\phi(\bm{x}_2),\ldots,\phi(\bm{x}_n)]\in\mathbb{R}^{l\times n}$.
We show this matrix can be approximated by a matrix of low rank\footnote{We present all the proofs in the supplementary material.}.
\begin{theorem}\label{theo_rank_phi}
Suppose $\bm{X}$ is given by Assumption \ref{asp_fz_analytic}. Then for any $q'$ obeying $\binom{d+q'}{q'}\leq \min\lbrace l,n\rbrace$, there exists a matrix $\bm{\Phi}$ with rank at most $\binom{d+q'}{q'}$, such that
$$\Vert\phi(\bm{X})-\bm{\Phi}\Vert_{\infty}\leq c_{q'},$$
where $c_{q'}=\tfrac{c_z^{q'+1}}{(q'+1)!}\max_{\Vert \bm{z}\Vert_{\infty} \leq c_z}\Vert(\phi\circ f)^{(q'+1)}(\bm{z})\Vert_\infty$.
\end{theorem}
Theorem \ref{theo_rank_phi} states that,
when $c_{q'}$ is small, $\phi(\bm{X})$ is approximately low-rank.
The value of $c_{q'}$ is related to the moduli of smoothness of $f$ and $\phi$,
which can be arbitrarily small when $f$ is sufficiently smooth
and $q'$ is sufficiently large.
For example, when $q'=1$, $\textup{rank}(\bm{\Phi})\leq d+1$ and $c_q'$ could be large.
When $q'=4$, $\textup{rank}(\bm{\Phi})\leq \binom{d+4}{4}$ and $c_q'$ is zero provided that the $5$-th order derivative of $\phi\circ f$ vanishes.

As a special case of Assumption \ref{asp_fz_analytic}, we introduce:
\begin{assumption}\label{asp_fz_poly}
$\bm{X}\in\mathbb{R}^{m\times n}$ is given by Assumption \ref{asp_fz_analytic}, in which $f$ consists of polynomials of order at most $\alpha$.
\end{assumption}
\noindent

For this special case, we have the following lemma:
\begin{lemma}\label{theo_rank_phi_poly}
Suppose $\bm{X}$ is given by Assumption \ref{asp_fz_poly}. Then
\begin{eqnarray*}
  \textup{rank}(\bm{X})&\leq&\min\lbrace \scaleobj{0.7}{\binom{d+\alpha}{\alpha}},m,n\rbrace, \\
  \textup{rank}(\phi(\bm{X}))&\leq&\min\lbrace \scaleobj{0.7}{\binom{d+\alpha q}{\alpha q},\binom{m+q}{q}},n\rbrace.
\end{eqnarray*}
\end{lemma}

Note that when $\alpha$ is large, the rank of $\bm{X}$ will be high or even full,
whereas when  $d\ll m$ and $n$ is sufficiently large, $\phi(\bm{X})$ is  low-rank.
Consider the following example.
\begin{example}
Let $m=10$, $d=2$, $n=50$, $\alpha=3$, and $q=2$. We have $\bm{X}\in\mathbb{R}^{10\times 50}$, $\phi(\bm{X})\in\mathbb{R}^{66\times 50}$, $\textup{rank}(\bm{X})=10$, and $\textup{rank}(\phi(\bm{X}))=28$.
\end{example}

Inspired by Lemma \ref{theo_rank_phi_poly}, for $\bm{X}$ given by Assumption \ref{asp_fz_poly},  we seek to recover the missing entries
by solving the \emph{polynomial matrix completion} (PMC) problem:
\begin{equation}\label{Eq.rankmin_PhiX}
\mathop{\text{minimize}}_{\hat{\bm{X}}}\ \textup{rank}(\phi(\hat{\bm{X}})),
\textup{subject to}\ \mathcal{P}_{\Omega}(\hat{\bm{X}})=\mathcal{P}_{\Omega}(\bm{X}).
\end{equation}
\begin{lemma}\label{theo_assump12}
For any $\bm{X}$ following Assumption \ref{asp_fz_analytic}
and any $\alpha \in \mathbb{N}$,
there is some $\tilde{\bm{X}}$ following Assumption \ref{asp_fz_poly} such that
$$\Vert\bm{X}-\tilde{\bm{X}}\Vert_{\infty}\leq c_{\alpha},$$
where $c_{\alpha}=\tfrac{c_z^{\alpha+1}}{(\alpha+1)!}\max_{\Vert \bm{z}\Vert_{\infty} \leq c_z}\Vert f^{(\alpha+1)}(\bm{z})\Vert_\infty$.
\end{lemma}
Lemma \ref{theo_assump12} suggests that we may approximate
any $\bm{X}$ that follows Assumption \ref{asp_fz_analytic} by
another matrix $\tilde{\bm{X}}$ that follows Assumption \ref{asp_fz_poly}
whenever $f$ is smooth enough.
By Lemma \ref{theo_rank_phi_poly}, the corresponding $\phi(\tilde{\bm{X}})$ is low-rank.
The phenomenon is consistent with Theorem \ref{theo_rank_phi}.
Hence to recover the missing entries of $\bm{X}$ following Assumption \ref{asp_fz_analytic}, we seek to solve
\begin{equation}\label{Eq.rankmin_PhiX_E}
\begin{aligned}
&\mathop{\text{minimize}}_{\hat{\bm{X}},\hat{\bm{E}}}\ \textup{rank}(\phi(\hat{\bm{X}}))+\tfrac{\lambda}{2}\Vert\hat{\bm{E}}\Vert_F^2,\\
&\textup{subject to}\ \mathcal{P}_{\Omega}(\hat{\bm{X}}+\hat{\bm{E}})=\mathcal{P}_{\Omega}(\bm{X}),
\end{aligned}
\end{equation}
where $\hat{\bm{E}}$ denotes small residuals, $\lambda$ is a regularization parameter, and the solution of $\hat{\bm{X}}$ is an estimation of $\tilde{\bm{X}}$ in Lemma \ref{theo_assump12}.
The formulation \eqref{Eq.rankmin_PhiX_E} is also useful when
the matrix $\bm{X}$ is generated from Assumption \ref{asp_fz_poly}
with some small additive noise.
In the remainder of this section,
we will suppose $\bm{X}$ is given by Assumption \ref{asp_fz_poly}
and focus on problem \eqref{Eq.rankmin_PhiX}.

To analyze the sample complexity for the matrix completion problem \eqref{Eq.rankmin_PhiX}, we present the following theorem:
\begin{theorem}\label{theo_mnp}
Let $\textup{mnp}_{\phi}(\bm{X})$ be the minimum number of parameters required to determine $\bm{X}$ uniquely among all matrices in the set $\lbrace \bm{\mathcal{X}}\in\mathbb{R}^{m\times n}:\ \textup{rank}(\phi(\bm{\mathcal{X}}))=\tilde{d}\rbrace$\footnote{It is worth noting that the minimum number of parameters required to determine $\bm{X}$ uniquely among all matrices defined by Assumption \ref{asp_fz_analytic} or \ref{asp_fz_poly} is much lower. However, in this paper, our method is based on the rank minimization for $\phi(\bm{X})$ but not explicit polynomial regression for $\bm{X}$.}. Define $\tilde{r}:=\min\lbrace o:\scaleobj{0.9}{\binom{o+q}{q}\geq \tilde{d}}\rbrace$. Then
\begin{equation}\label{Eq.bound_HRMC}
\textup{mnp}_{\phi}(\bm{X})=(m-\tilde{r}) \tilde{d}+n\tilde{r}.
\end{equation}
\end{theorem}
\begin{remark}
\textup{
It is easy to show that $\big(q!\times\tilde{d}\big)^{1/q}-q\leq \tilde{r}\leq \big(q!\times\tilde{d}\big)^{1/q}$. Thus we can compute $\tilde{r}$ via at most $q+1$ trials.}
\end{remark}

Compared to Theorem \ref{theo_mnp}, the minimum number of parameters (number of degrees of freedom) required to determine a rank-$r$ matrix $\bm{X}$ uniquely is
\begin{equation}
\textup{mnp}(\bm{X}):=(m-r)r+nr.
\end{equation}
When $n\geq m$, the number of samples $\Omega$ required for LRMC methods to succeed \cite{CandesRecht2009} is
\begin{equation}\label{Eq.bound_LRMC}
\vert \Omega\vert\geq C_1 nr\log n,
\end{equation}
where $r=\binom{d+p}{p}$ and $C_1$ is a numerical constant independent of $r$ and $n$.  Here the $\log n$ factor accounts for the coupon collecter effect to ensure all columns and rows of the matrix are sampled with high probability \cite{CandesRecht2009}.
Similarly, we suspect that Problem~\eqref{Eq.rankmin_PhiX} cannot recover $\bm{X}$ unless
\begin{equation}\label{Eq.bound_PMC}
\vert \Omega\vert\geq C_2 n\tilde{r}\log n,
\end{equation}
where $C_2$ is a numerical constant, provided that $n\geq \tfrac{(m-\tilde{r}) \tilde{d}}{\tilde{r}}$.
Note that $\tilde{r}$ in (\ref{Eq.bound_PMC}) is much smaller than $r$ in (\ref{Eq.bound_LRMC}).
\begin{example}
Without loss of generality, we assume $\textup{rank}(\phi(\bm{X}))=\tilde{d}=\binom{d+\alpha q}{\alpha q}$. When $d=3$, $p=2$, and $q=3$, we have $r=10$ and $\tilde{r}=6$.
\end{example}
\noindent
In other words, to recover the missing entries of $\bm{X}$, rank minimization on $\phi(\bm{X})$ requires fewer observed entries than rank minimization on $\bm{X}$ does.
Our numerical results confirm this observation.

\section{Tractable Relaxations of PMC}
\subsection{Rank Relaxations in Feature Space}
As rank minimization is NP-hard, we suggest solving (\ref{Eq.rankmin_PhiX}) by relaxing the problem as
\begin{equation}\label{Eq.rankmin_PhiX_relaxation}
\mathop{\text{minimize}}_{\hat{\bm{X}}}\ \mathcal{R}(\phi(\hat{\bm{X}})),
\textup{subject to}\ \mathcal{P}_{\Omega}(\hat{\bm{X}})=\mathcal{P}_{\Omega}(\bm{X}),
\end{equation}
where $\mathcal{R}(\phi(\hat{\bm{X}}))$ is a relaxation of $\textup{rank}(\phi(\hat{\bm{X}}))$.
A standard choice of $\mathcal{R}(\phi(\hat{\bm{X}}))$ is the nuclear norm penalty $\Vert \phi(\hat{\bm{X}})\Vert_\ast$ .
More generally,
we may use a Schatten-$p$ (quasi) norm ($0<p\leq 1$) \cite{Nie2015} to relax $\textup{rank}(\phi(\hat{\bm{X}}))$:
\begin{equation}
\mathcal{R}_1(\phi(\hat{\bm{X}})):=\Vert \phi(\hat{\bm{X}})\Vert_{S_p}^p=\sum_{i=1}^n\sigma_i^{p}(\phi(\hat{\bm{X}})).
\end{equation}
Notice that the Schatten-$1$ norm is the nuclear norm.

Since $\mathcal{R}_1(\phi(\hat{\bm{X}}))$ penalizes all the singular values of $\phi(\hat{\bm{X}})$, it is very different from exact rank minimization for $\phi(\hat{\bm{X}})$ (unless $p\rightarrow 0$). Inspired by \cite{MC_TNNR_PAMI2013,TIP_MC_TNNM}, in this paper, we propose to regularize only the smallest singular values:
\begin{equation}
\mathcal{R}_2(\phi(\hat{\bm{X}})):=\Vert \phi(\hat{\bm{X}})\Vert_{S_p\vert s}^p=\sum_{i=s+1}^n\sigma_i^{p}(\phi(\hat{\bm{X}})),
\end{equation}
where $0<p\leq 1$, $\sigma_1\geq\sigma_2\geq\cdots\geq\sigma_n$, and $s\leq \textup{rank}(\phi(\hat{\bm{X}}))$. Minimizing $\mathcal{R}_2(\phi(\hat{\bm{X}}))$ will only minimize the small singular values of $\phi(\hat{\bm{X}})$.
This approach can improve upon standard Schatten-$p$ norm penalization
when we know a good lower bound on the rank.
For the PMC problem, 
we suggest setting $s=m$,
since $\textup{rank}(\phi(\bm{X}))\geq \textup{rank}(\bm{X})=m$ by Assumption \ref{asp_fz_poly} (when $\alpha$ is large).

We can further improve our results by applying more strict scrutiny
to small singular values
\cite{xie2016weighted}.
Concretely, we propose the following relaxation of $\textup{rank}(\phi(\hat{\bm{X}}))$:
\begin{equation}
\mathcal{R}_3(\phi(\hat{\bm{X}})):=\Vert \phi(\hat{\bm{X}})\Vert_{S_p\vert \bm{w}}^p=\sum_{i=1}^nw_i \sigma_i^{p}(\phi(\hat{\bm{X}})),
\end{equation}
where the weights
$w_1\leq w_2\leq\cdots \leq w_n$ are increasing.
Notice that $\mathcal{R}_3(\phi(\hat{\bm{X}}))$ is identical to $\mathcal{R}_2(\phi(\hat{\bm{X}}))$ if $w_1=\cdots=w_s=0$ and $w_{s+1}=\cdots=w_n=1$.

One straightforward choice,
inspired by the log-det heuristic for $\ell_0$ norm minimization \cite{fazel2003log},
sets $\bm{w}_1:=[1/(\sigma_1^p+\epsilon),\ldots,1/(\sigma_n^p+\epsilon)]^T$.
Here $\bm{\sigma}$ are the singular values of an initial approximation to $\bm{X}$
obtained by solving Problem \label{Eq.rankmin_PhiX_relaxation} with Schatten-$p$ norm regularization, and
$\epsilon$ is a small constant chosen for numerical stability.
Using these weights $\bm{w}$ focuses the optimization on pushing the
the smaller singular values of $\phi(\hat{\bm{X})}$ towards zero.
These smaller singular values add to the rank, but contribute negligibly to the overall approximation quality.
In addition, we found that the simpler heuristic $\bm{w}_2:=[1/n,2/n,\ldots,1]^T$
performed nearly as well as $\bm{w}_1$ in our numerical results.

\subsection{Kernel Representation}
When using $\mathcal{R}_1(\phi(\hat{\bm{X}}))$, it is possible to solve (\ref{Eq.rankmin_PhiX_relaxation}) directly. But the computation cost is very high and the optimization is difficult when $q$ and $m$ are not small. As $\Vert \phi(\bm{X})\Vert_{S_p}=\big(\textup{Tr}((\phi(\hat{\bm{X}})^T\phi(\hat{\bm{X}}))^{p/2})\big)^{1/p}$, we have
\begin{equation}\label{Eq.Spmin_trace}
\mathcal{R}_1(\phi(\hat{\bm{X}}))=\textup{Tr}\Big(\mathcal{K}(\hat{\bm{X}})^{p/2}\Big),
\end{equation}
in which we have replaced the $n\times n$ Gram matrix $\phi(\hat{\bm{X}})^T\phi(\hat{\bm{X}})$ with a kernel matrix $\mathcal{K}(\hat{\bm{X}})$, i.e., for $i,j=1,2,\ldots,n$
$$[\mathcal{K}(\hat{\bm{X}})]_{ij}=\phi(\hat{\bm{x}}_i)^T\phi(\hat{\bm{x}}_j)=k(\hat{\bm{x}}_i,\hat{\bm{x}}_j),
$$
where $k(\cdot,\cdot)$ denotes a kernel function. Hence, we obtain $\mathcal{R}_1(\phi(\hat{\bm{X}}))$ without explicitly carrying out the nonlinear mapping $\phi$ provided that the corresponding kernel function exists. The existence, choice, and  property of kernel function will be detailed later. We first illustrate the kernel representations of $\mathcal{R}_2(\phi(\hat{\bm{X}}))$ and $\mathcal{R}_3(\phi(\hat{\bm{X}}))$, which are not straightforward, compared to that of $\mathcal{R}_1(\phi(\hat{\bm{X}}))$.

The following lemma enables us to kernelize $\mathcal{R}_2(\phi(\hat{\bm{X}}))$:
\begin{lemma}\label{lem_ts}
For any $\bm{X}\in\mathbb{R}^{m\times n}$,
\begin{equation*}
\begin{aligned}
\Vert \phi(\bm{X})\Vert_{S_p\vert s}^p=&\textup{Tr}(\mathcal{K}(\bm{X})^{p/2})\\
&-\max_{\bm{P}^T\bm{P}=\bm{I}_s,\bm{P}\in\mathbb{R}^{n\times s}}\textup{Tr}\Big(\big(\bm{P}^T\mathcal{K}(\bm{X})\bm{P}\big)^{p/2}\Big).
\end{aligned}
\end{equation*}
\end{lemma}
\begin{remark}
\textup{It is worth mentioning that, for any fixed $\bm{X}$, $\max_{\scaleobj{0.8}{\bm{P}^T\bm{P}=\bm{I}_s}}\textup{Tr}\scaleobj{0.9}{\big(\big(\bm{P}^T\mathcal{K}(\bm{X})\bm{P}\big)^{p/2}\big)}=\textup{Tr}\big(\big(\bm{V}_s^T\mathcal{K}(\bm{X})\bm{V}_s\big)^{p/2}\big)$, where $\bm{V}_s$ consists of the eigenvectors of $\mathcal{K}(\bm{X})$ corresponding to the largest $s$ eigenvalues. When $\bm{X}$ is an unknown variable, the formulation of $\Vert \phi(\bm{X})\Vert_{S_p\vert s}^p$ in terms of $\bm{V}_s$ is not applicable.}
\end{remark}

The following lemma enables us to kernelize $\mathcal{R}_3(\phi(\hat{\bm{X}}))$:
\begin{lemma}\label{lem_ws}
For any $\bm{X}\in\mathbb{R}^{m\times n}$,
$$
\scaleobj{0.9}{\Vert \phi(\bm{X})\Vert_{S_p\vert \bm{w}}^p=\min_{\bm{Q}^T\bm{Q}=\bm{Q}\bm{Q}^T=\bm{I}_n}\textup{Tr}\Big(\big(\bm{W}^{1/p}\bm{Q}^T\mathcal{K}(\bm{X})\bm{Q}\bm{W}^{1/p}\big)^{p/2}\Big)}.
$$
\end{lemma}
\begin{remark}
\textup{For any fixed $\bm{X}$, $\Vert \phi(\bm{X})\Vert_{S_p\vert \bm{w}}^p=\textup{Tr}\big(\bm{W}\bm{V}^T\mathcal{K}(\bm{X})^{p/2}\bm{V}\big)$, where $\bm{V}$ denotes the eigenvectors of $\mathcal{K}(\bm{X})$ corresponding to the eigenvalues in descending order. When $\bm{X}$ is an unknown variable, the formulation of $\Vert \phi(\bm{X})\Vert_{S_p\vert \bm{w}}^p$ in terms of $\bm{V}$ is not applicable.}
\end{remark}

As we have provided the kernel representations of $\lbrace \mathcal{R}_j(\phi(\hat{\bm{X}}))\rbrace_{j=1}^3$, now let us detail the choice of kernel functions and analyze the corresponding properties. The previous analysis such as Theorem \ref{theo_rank_phi}, Lemma \ref{theo_rank_phi_poly}, and the formulation (\ref{Eq.rankmin_PhiX_relaxation}) are applicable to any polynomial feature map $\phi$ given by Definition \ref{def_phi}. Hence, we only need to ensure that the selected kernel functions induce polynomial feature maps. First, consider the $a$-order polynomial function kernel
$$k^{\scaleobj{0.7}{poly}}(\bm{x},\bm{y})=\left(\bm{x}^T\bm{y}+b\right)^a$$
where $b$ is a parameter trading off the influence of higher-order versus lower-order terms in the polynomial. The corresponding $\phi$ is an $a$-order polynomial feature map \cite{KPCA1998}. Moreover, for any $\phi$ given by Definition \ref{def_phi}, we can always construct a kernel accordingly, via combining different polynomial kernels linearly.
Consider the (Gaussian) radial basis function (RBF)
$$
k^{\scaleobj{0.6}{RBF}}(\bm{x},\bm{y})=\textup{exp}\left(-\Vert \bm{x}-\bm{y}\Vert^2/(2\sigma^2)\right),
$$
where the hyper-parameter $\sigma$ controls the smoothness of the kernel. The $\phi$ implicitly determined by RBF kernel is an infinite-order polynomial feature map because RBF kernel is a weighted sum of polynomial kernels of orders from $0$ to $\infty$.
Using Lemma \ref{theo_rank_phi_poly}, we obtain
\begin{corollary}\label{theo_rank_phi_polyrbf}
Let $\phi_q(\bm{x})$ and $\phi_{\sigma}(\bm{x})$ be the feature maps of $q$-order polynomial kernel and RBF kernel with parameter $\sigma$ respectively. Define $\phi_{\lbrace\leq q\rbrace}(\bm{x}):=\lbrace\tfrac{c_x}{\sigma^{2j}j!}\phi_j(\bm{x})\rbrace_{0\leq j\leq q}$, where $c_x=\exp(-\tfrac{\Vert \bm{x}\Vert_2^2}{\sigma^2})$.
Suppose $\bm{X}$ is given by Assumption \ref{asp_fz_poly} and $\lbrace l,n\rbrace$ are sufficiently large. Then for any $q$,
$$
\textup{rank}(\phi_q(\bm{X}))=\textup{rank}(\phi_{\lbrace\leq q\rbrace}(\bm{X}))\leq\scaleobj{0.7}{\binom{d+\alpha q}{\alpha q}},
$$
and
$$
\phi_{\sigma}(\bm{X})^T\phi_{\sigma}(\bm{X})=\phi_{\lbrace\leq q\rbrace}(\bm{X})^T\phi_{\lbrace\leq q\rbrace}(\bm{X})+\bm{E}_{q\sigma},
$$
where $\bm{E}_{q\sigma}\in\mathbb{R}^{n\times n}$ and $\vert[\bm{E}_{q\sigma}]_{ij}\vert\leq \exp(-\tfrac{\min_{j}\Vert \bm{x}_j\Vert_2^2}{\sigma^2})\tfrac{\max_{j}\Vert \bm{x}_j\Vert_2^{q+1}}{\sigma^{2(q+1)}(q+1)!}$.
\end{corollary}
Corollary \ref{theo_rank_phi_polyrbf} indicates that when we use polynomial kernel of relatively low order, $\phi(\bm{X})$ is exactly of low-rank, provided that $l$ and $n$ are sufficiently large. When we use RBF kernel, $\phi(\bm{X})$ can be well approximated by that of sum of polynomial kernels, provided that $\sigma$ is large enough.

It is worth noting that the regularization $\mathcal{R}_1(\phi(\hat{\bm{X}}))$ has been utilized in \cite{pmlr-v70-ongie17a,FANNLMC} but the data generation model and rank property are different.
For example, the assumption in \cite{pmlr-v70-ongie17a} is based on algebraic variety and the ranks of $\bm{X}$ and $\phi(\bm{X})$ cannot be explicitly computed. In our paper, the assumption is based on analytic function, which differs from algebraic variety and enables us to obtain the ranks of $\bm{X}$ and $\phi(\bm{X})$ explicitly (e.g. Lemma \ref{theo_rank_phi_poly}). Moreover, $\mathcal{R}_2(\phi(\hat{\bm{X}}))$ and $\mathcal{R}_3(\phi(\hat{\bm{X}}))$ are able to outperform $\mathcal{R}_1(\phi(\hat{\bm{X}}))$, which will be validated in the experiments.

\section{Optimization for PMC}
Solving (\ref{Eq.rankmin_PhiX_relaxation}) especially with $\mathcal{R}_2(\phi(\hat{\bm{X}}))$ and $\mathcal{R}_3(\phi(\hat{\bm{X}}))$ are challenging due to the nonconvexity of kernel function and the presence of $\bm{P}$ and $\bm{Q}$ in the objective functions. However, Lemma \ref{lem_ts} and Lemma \ref{lem_ws} provided the upper bounds of $\mathcal{R}_2(\phi(\hat{\bm{X}}))$ and $\mathcal{R}_3(\phi(\hat{\bm{X}}))$.
When using $\mathcal{R}_2(\phi(\hat{\bm{X}}))$, we propose to solve the following two problems alternately:
\begin{equation}\label{Eq.AM_S}
\begin{aligned}
&\bm{P}_t=\quad \mathop{\arg\max}_{\bm{P}^T\bm{P}=\bm{I}_s}\quad\ \scaleobj{0.9}{\textup{Tr}\Big(\big(\bm{P}^T\mathcal{K}(\hat{\bm{X}}_{t-1})\bm{P}\big)^{p/2}\Big)=\bm{V}_s},\\
&\hat{\bm{X}}_t=\mathop{\arg\min}_{\scaleobj{0.9}{\mathcal{P}_{\Omega}(\hat{\bm{X}})=\mathcal{P}_{\Omega}(\bm{X})}} \scaleobj{0.9}{\textup{Tr}\big(\mathcal{K}(\hat{\bm{X}})^{p/2}\big)-\textup{Tr}\Big(\big(\bm{P}_t^T\mathcal{K}(\hat{\bm{X}})\bm{P}_t\big)^{p/2}\Big)},
\end{aligned}
\end{equation}
where $\bm{V}_s$ denotes the eigenvectors of $\mathcal{K}(\hat{\bm{X}}_{t-1})$ corresponding to the largest $s$ eigenvalues.  
When using $\mathcal{R}_3(\phi(\hat{\bm{X}}))$, we propose solve the following two problems alternately:
\begin{equation}\label{Eq.AM_W}
\begin{aligned}
&\bm{Q}_t=\mathop{\arg\min}_{\scaleobj{0.8}{\bm{Q}^T\bm{Q}=\bm{Q}\bm{Q}^T=\bm{I}_n}}\ \scaleobj{0.9}{\textup{Tr}\Big(\big(\bm{W}^{\tfrac{1}{p}}\bm{Q}^T\mathcal{K}(\hat{\bm{X}}_{t-1})\bm{Q}\bm{W}^{\tfrac{1}{p}}\big)^{p/2}\Big)}=\scaleobj{0.9}{\bm{V}},\\
&\hat{\bm{X}}_t=\mathop{\arg\min}_{\mathcal{P}_{\Omega}(\hat{\bm{X}})=\mathcal{P}_{\Omega}(\bm{X})} \textup{Tr}\Big(\big(\bm{W}^{\tfrac{1}{p}}\bm{Q}_t^T\mathcal{K}(\hat{\bm{X}})\bm{Q}_t\bm{W}^{\tfrac{1}{p}}\big)^{p/2}\Big),
\end{aligned}
\end{equation}
where $\bm{V}$ denotes the eigenvectors of $\mathcal{K}(\hat{\bm{X}}_{t-1})$ corresponding to the eigenvalues in descending order. 

We unify \eqref{Eq.AM_S} and \eqref{Eq.AM_W} into
\begin{equation}\label{Eq.AM_unif}
\hat{\bm{X}}_t=\mathop{\arg\min}_{\mathcal{P}_{\Omega}(\hat{\bm{X}})=\mathcal{P}_{\Omega}(\bm{X})}\ \mathcal{L}(\hat{\bm{X}},\bm{\Theta}_{t-1}),
\end{equation}
where the $\bm{\Theta}_{t-1}=\bm{P}_t$ or $\bm{Q}_t\bm{W}^{1/p}$. For a given $\bm{\Theta}_{t-1}$, there is no need to compute $\hat{\bm{X}}_t$ exactly. We propose to update the unknown entries of $\hat{\bm{X}}$ (denoted by $[\hat{\bm{X}}]_{\bar{\Omega}}$) via
\begin{equation}\label{Eq.AM_gd}
[\hat{\bm{X}}_t]_{\bar{\Omega}}\leftarrow [\hat{\bm{X}}_{t-1}]_{\bar{\Omega}}-\mu_t[\nabla_{\hat{\bm{X}}}\mathcal{L}(\hat{\bm{X}},\bm{\Theta}_{t-1})]_{\bar{\Omega}},
\end{equation}
where $\mu_t$ is the step size, $\bm{\Theta}_{t-1}$ is estimated from $\hat{\bm{X}}_{t-1}$, and the gradient $\nabla_{\hat{\bm{X}}}\mathcal{L}$ is computed through using chain rule. The convergence speed of naive gradient descent is low. We can use second order methods (e.g. quasi-Newton methods) to accelerate the optimization, which, however, have high computational costs when the number of unknowns is large.

One popular method of gradient descent optimization is the Adam algorithm \cite{adam2014}, which has achieved considerable success in deep learning. 
Adam requires determining the step size beforehand. We propose to adaptively tune the step size, which yields a variant of Adam we call Adam+. The details of performing \eqref{Eq.AM_gd} via Adam+ are shown in Algorithm \ref{alg.Adam+}. The space complexity and time complexity (per iteration) of our method PMC are $O(n^2)$ and $O(n^3)$, which are nearly the same as those of VMC \cite{pmlr-v70-ongie17a} and NLMC \cite{FANNLMC}. When we perform partial SVD algorithm (e.g. randomized SVD \cite{randomsvd}) in the optimization, the time complexity becomes $O(rn^2)$, where $r$ is the approximate rank of $\phi(\bm{X})$ and $r<<n$.

\renewcommand{\algorithmicrequire}{\textbf{Input:}}
\renewcommand{\algorithmicensure}{\textbf{Output:}}
\begin{algorithm}[h]
\caption{Optimization for PMC using Adam+}
\label{alg.Adam+}
\begin{algorithmic}[1]
\Require
$\bm{X}$, $\Omega$, $k(\cdot,\cdot)$, $s$ or $\bm{w}$, $\gamma=10^{-6}$, $\lambda=10^{-4}$, $\varepsilon=10^{-6}$, $t_{max}$, $\beta_1=0.9$, $\beta_2=0.999$, $\epsilon=10^{-8}$, $t=0$
\State \textbf{initialize} $[\hat{\bm{X}}]_{\bar{\Omega}}=0$, $\hat{\bm{x}}=\textup{vec}([\hat{\bm{X}}]_{\bar{\Omega}})$, $\bm{m}_0=\bm{\upsilon}_0=\bm{0}$
\Repeat
\State $t\leftarrow t+1$
\State perform SVD: $\mathcal{K}(\hat{\bm{X}})=\bm{V}\bm{S}\bm{V}^T$
\State compute $\bm{\Theta}_{t-1}$ and $\nabla_{\hat{\bm{X}}}\mathcal{L}(\hat{\bm{X}},\bm{\Theta}_{t-1})$
\State $\bm{g}_t\leftarrow \textup{vec}\big([\nabla_{\hat{\bm{X}}}\mathcal{L}(\hat{\bm{X}},\bm{\Theta}_{t-1})]_{\bar{\Omega}}\big)$
\State $\bm{m}_t\leftarrow \beta_1\bm{m}_{t-1}+(1-\beta_1)\bm{g}_t$
\State $\bm{\upsilon}_t\leftarrow \beta_2\bm{\upsilon}_{t-1}+(1-\beta_2)\bm{g}_t^2$
\State $\hat{\bm{m}}_t\leftarrow \bm{m}/(1-\beta_1^t)$;\quad $\hat{\bm{\upsilon}}_t\leftarrow \bm{\upsilon}/(1-\beta_2^t)$
\State $\hat{\bm{x}}_t\leftarrow \hat{\bm{x}}_{t-1}-\lambda\hat{\bm{m}}_t/(\sqrt{\hat{\bm{\upsilon}}_t}+\epsilon)$;\quad $[\hat{\bm{X}}]_{\bar{\Omega}}\leftarrow \hat{\bm{x}}_t$
\State \textbf{if} $\mathcal{L}_t>\mathcal{L}_{t-1}$
\State \quad $\lambda\leftarrow 0.8\lambda$ \quad \textbf{else} \quad $\lambda\leftarrow 1.1\lambda$
\State \textbf{endif}
\Until{$\vert \hat{\bm{x}}_t-\hat{\bm{x}}_{t-1}\vert_{\infty}<\varepsilon$ or $t=t_{max}$}
\Ensure $\hat{\bm{X}}$
\end{algorithmic}
\end{algorithm}

\section{Generalization for multiple manifolds}\label{sec_mm}
More generally, the columns of $\bm{X}$ can be drawn from different manifolds. We then extend Assumption \ref{asp_fz_poly} to
\begin{assumption}\label{def_XUoS}
Suppose $\bm{X}$ consists of the columns of $[\bm{X}^{\lbrace 1\rbrace},\bm{X}^{\lbrace 2\rbrace},\ldots,\bm{X}^{\lbrace k\rbrace}]$ and for all $j=1,\ldots, k$, $\bm{X}^{\lbrace j\rbrace}\in\mathbb{R}^{m\times n_j}$ satisfies Assumption \ref{asp_fz_poly} with different $f^{\lbrace j\rbrace}$.
\end{assumption}
Without loss of generality, we let $d_1=\cdots=d_k=d$ and $n_1=\cdots=n_k=n$.
Corollary \ref{theo_rank_phi_polyrbf} can be easily extended to

\begin{corollary}\label{theo_rank_phi_polyrbf_uos}
Use the same notations in Corollary \ref{theo_rank_phi_polyrbf}. Suppose $\bm{X}$ is given by Assumption \ref{def_XUoS} and $\lbrace l,n\rbrace$ are sufficiently large. Then for any $q$,
$$
\textup{rank}(\phi_q(\bm{X}))=\textup{rank}(\phi_{\lbrace\leq q\rbrace}(\bm{X}))\leq k\scaleobj{0.8}{\binom{d+\alpha q}{\alpha q}},$$
and $$
\phi_{\sigma}(\bm{X})^T\phi_{\sigma}(\bm{X})=\phi_{\lbrace\leq q\rbrace}(\bm{X})^T\phi_{\lbrace\leq q\rbrace}(\bm{X})+\bm{E}_{q\sigma},
$$
where $\bm{E}_{q\sigma}\in\mathbb{R}^{kn\times kn}$ and $\vert[\bm{E}_{q\sigma}]_{ij}\vert\leq e^{-\tfrac{\min_{j}\Vert \bm{x}_j\Vert_2^2}{\sigma^2}}\tfrac{\max_{j}\Vert \bm{x}_j\Vert_2^{q+1}}{\sigma^{2(q+1)}(q+1)!}$.
\end{corollary}
\noindent Consequently, it is possible to recover $\bm{X}$ through minimizing the rank of $\phi(\bm{X})$. Moreover, let $r\leq k\binom{d+p}{p}$ and $\tilde{d}\leq k\binom{d+\alpha q}{\alpha q}$, then Theorem \ref{theo_mnp} and bounds (\ref{Eq.bound_LRMC}) and (\ref{Eq.bound_PMC}) are also applicable in this case. The following example indicates that rank minimization on $\phi(\bm{X})$ requires fewer observed entries, compared to rank minimization on $\bm{X}$.
\begin{example}
When $d=3$, $p=2$, $k=3$, and $q=3$, we have $r=30$ and $\tilde{r}=10$.
\end{example}
\noindent

\section{Transductive learning}
In transductive learning, the testing data is used in the training step. We can use matrix completion to do classification. Specifically, let $\lbrace \bm{X}, \bm{Y}\rbrace$ be the training data, where $\bm{X}\in\mathbb{R}^{m\times n}$ is the feature matrix, $\bm{Y}\in\mathbb{R}^{c\times n}$ is the one-hot label matrix, and $c$ is the number of classes. Let $\lbrace \bm{X}', \bm{Y}'\rbrace$ be the testing data, where $\bm{X}'\in\mathbb{R}^{m\times n'}$ is the feature matrix, $\bm{Y}'\in\mathbb{R}^{c\times n'}$ is the unknown label matrix. First, consider a linear classifier $\bm{y}=\bm{W}\bm{x}+\bm{b}$ and let
\begin{equation}
\overline{\bm{X}}=\left[
\begin{matrix}
\bm{X}&\bm{X}'\\
\bm{Y}&\bm{Y}'\\
\end{matrix}
\right].
\end{equation}
We can regard $\bm{Y}'$ as the missing entries of $\overline{\bm{X}}$ and perform LRMC methods to obtain $\bm{Y}'$. It will be more useful when $\bm{X}$, $\bm{X}'$, or/and $\bm{Y}$ have missing entries. However, we cannot obtain nonlinear classification via LRMC methods. In addition, when the number of classes is large, the rank of $\overline{\bm{X}}$ will be high, which is a challenge to LRMC methods.

In multi-class classification, the data are often drawn from multiple nonlinear manifolds and one manifold corresponds to one class:
\begin{equation}\label{Eq.nl_class_00}
\bm{x}_i=f^{\lbrace j\rbrace}(\bm{z}_i),\quad \textup{if}\ \bm{x}_i \in C_j,
\end{equation}
where $C_j$ denotes the class $j$.
For each class $C_k$, there exists a $h_k:\mathbb{R}^m\rightarrow \mathbb{R}^c$ that maps the feature to the label:
\begin{equation}\label{Eq.nl_class_0}
\bm{y}_i=h_j(\bm{x}_j)=h_j\big(f^{\lbrace j\rbrace}(\bm{z}_i)\big),\quad \textup{if}\ \bm{x}_i \in C_j.
\end{equation}
In practice, it is difficult to obtain $\lbrace h_j\rbrace_{j=1}^c$ separately. For example, in linear regression or multi-layer neural network, we usually train a single $h$ to fit all data:
\begin{equation}
\bm{y}_i=h(\bm{x}_i),\quad \textup{if}\ \bm{x}_i \in C_j.
\end{equation}
In support vector machines, we usually train $c-1$ classifiers and each of them fit all data individually.

As (\ref{Eq.nl_class_00}) and (\ref{Eq.nl_class_0}) matches our Assumption \ref{def_XUoS} (let $k=c$), it is possible to construct  $\lbrace h_j\rbrace_{j=1}^c$ separately and implicitly if we perform PMC on $\overline{\bm{X}}$, even if $\overline{\bm{X}}$ is high-rank and highly incomplete. Therefore, the missing entries of $\overline{\bm{X}}$ (including the unknown labels $\bm{Y}'$) can be recovered by our PMC.

\begin{figure}[h!]
\centering
\includegraphics[width=7cm]{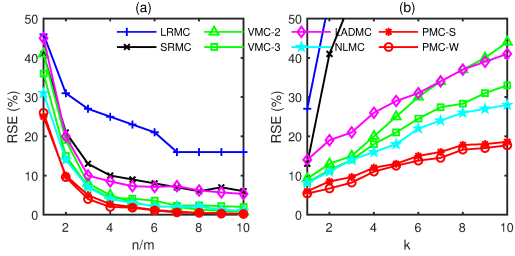}
\caption{(a) RSE on single-manifold data with different number of columns ($\rho=0.5$). (b) RSE on multiple-manifold data with different number of manifolds ($\rho=0.5$).}
\label{fig.1ks_50}
\end{figure}

\begin{figure}[t]
\includegraphics[width=8.5cm]{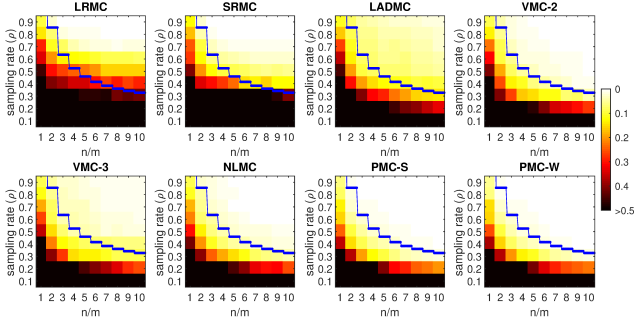}
\centering
\caption{RSE on single-manifold data with different sampling rate and different number of columns.}
\label{fig.1s}
\end{figure}

\section{Experiments}
We compare the proposed PMC methods with LRMC method (nulcear norm minimization), SRMC \cite{Fan2017290}, LADMC \cite{LADMC} (the iterative algorithm), VMC-2 (2-order polynomial kernel), VMC-3 (3-order polynomial kernel)\cite{pmlr-v70-ongie17a}, and NLMC \cite{FANNLMC} (RBF kernel) in matrix completion on synthetic data, subspace clustering on incomplete data, motion capture data recovery, and classification on incomplete data. Note that PMC with $\mathcal{R}_1$ is equivalent to the NLMC method of \cite{FANNLMC}. For convenience, the PMC methods with $\mathcal{R}_2$ and $\mathcal{R}_3$ are denoted as PMC-S and PMC-W respectively. More details about the parameter setting and dataset description are in the supplementary material.
\subsection{Synthetic data}\label{sec.syn}
We use the following polynomial mapping $f:\mathbb{R}^{d}\rightarrow\mathbb{R}^{m}$ to generate random matrices :
\begin{equation*}\label{Eq.Syn1}
\bm{X}=f(\bm{Z})\triangleq \bm{A}\bm{Z}+\tfrac{1}{2}(\bm{B}\bm{Z}^{\odot 2}+\bm{C}\bm{Z}^{\odot 3}+\bm{C}\bm{Z}^{\odot 4}),
\end{equation*}
where $\bm{A},\bm{B},\bm{C},\bm{D}\in\mathbb{R}^{m\times d}$ and $\bm{Z}\in \mathbb{R}^{d\times n}$. The entries of $\bm{A},\bm{B},\bm{C}$, and $\bm{D}$ are randomly drawn from $\mathcal{N}(0,1)$. The entries of $\bm{Z}$ are randomly drawn from $\mathcal{U}(-1,1)$ and $\bm{Z}^{\odot c}$ denotes the $c$-th power performed on all elements of $\bm{Z}$. We randomly sample a fraction (denoted by $\rho$) of the entries and use matrix completion methods to recover unknown entries. The performances of the methods are evaluated by the relative squared error $\textup{RSE}:=\sqrt{\sum_{(i,j)\in{\bar{\Omega}}}(\bm{X}_{ij}-\bm{\hat{X}}_{ij})^2/\sum_{(i,j)\in{\bar{\Omega}}}\bm{X}_{ij}^2}$,
where $\hat{\bm{X}}$ denotes the recovered matrix and $\bar{\Omega}$ consists of the positions of unknown entries of $\bm{X}$. We report the average RSE of 50 repeated trials.

\begin{figure}[h!]
\centering
\includegraphics[width=8.5cm]{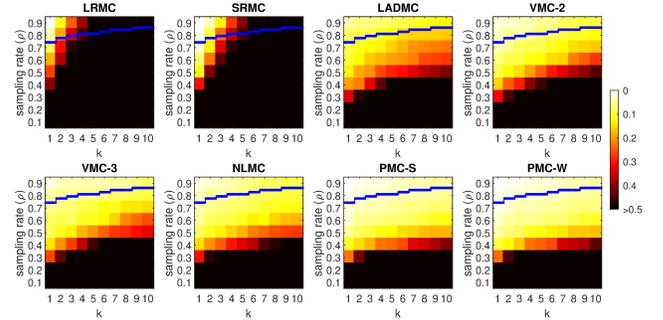}
\caption{RSE on multiple-manifold data with different sampling rate and different number of manifolds.}
\label{fig.ks}
\end{figure}

First, we set $d=2$ and $m=20$ and increase $n$ from $1m$ to $10m$. In these cases, the rank of $\bm{X}$ is $8$ although the intrinsic dimension is only $2$. The RSEs of all methods when $\rho=0.5$ are compared in Figure \ref{fig.1ks_50}(a). The recovery errors of VMC, NLMC, and our PMC decrease quickly when $n$ increases. Figure \ref{fig.1s} shows the RSEs of the eight methods in the cases of different number of columns $n$ and different sampling rate $\rho$. We see that PMC-S and PMC-W outperformed other methods. In Figure \ref{fig.1s}, the blue curve denotes the lower-bound of sampling rate we estimated with $\textup{mnp}_{\phi}(\bm{X})$ defined in (\ref{Eq.bound_HRMC}). Specifically, we set $q=3$ and have $\tilde{d}=73$. Then $\tilde{r}=6$ and the lower-bound of sampling rate is estimated as $\tfrac{(m-\tilde{r})\times \tilde{d}+n\tilde{r}}{mn}$. The performances of PMC-S and PMC-W are in according with the bound (\ref{Eq.bound_PMC}).

We set $d=2$, $m=20$, $n=50$ and use $k$ different $f$ to generate $k$ matrices of $m\times n$ to form a large matrix $\bm{X}\in\mathbb{R}^{m\times kn}$. Such $\bm{X}$ consists of data drawn from multiple manifolds and $\textup{rank}(\bm{X})=\min\lbrace 20,8k\rbrace$. The RSEs of all methods when $\rho=0.5$ are compared in Figure \ref{fig.1ks_50}(b). LRMC and SRMC fail when $k$ increases. The RSEs of PMC-S and PMC-W are much lower than those of other methods. Figure \ref{fig.ks} shows the RSEs of all methods in the cases of different $k$ and different $\rho$. We also estimated the lower bound (blue curve) of sampling rate using (\ref{Eq.bound_HRMC}). PMC-S and PMC-W always outperform other methods and comply with the bound. In addition, PMC-W is a little bit better than PMC-S.

\begin{table*}
\centering
\small
\caption{Classification errors ($\%$) on datasets with missing values}\label{Tab_class_results}
\begin{tabular}{ccccccccccc}
\hline
{Data set}	&	{$\theta$}	 & SVM & LRMC & {\scriptsize LRMC+SVM} & SRMC & VMC-2 	& VMC-3 & NLMC & PMC-S & PMC-W \\ \hline
\multirow{2}{*}{Mice protein}&10\% 	&8.96 	&6.33	&0.8	&2.96  &0.54	&0.46	&0.44	&0.41	&\textbf{0.39}		\\
						&50\% 		&32.5	&18.78	&5.26	&13.19  &1.24	&0.87	&0.81	&0.71	&\textbf{0.63}	\\ \hline

\multirow{2}{*}{Shuttle}&10\% 		&12.18 	&24.7	&\textbf{2.48}	&21.06 &9.7		&7.72	&4.8	&2.66		&3.86		\\
						&50\% 		&17.82	&28.7	&10.6		&27.3  &13.4	&11.1	&9.58	&\textbf{8.02}	&9.16	\\ \hline

\multirow{2}{*}{Dermatology}&10\% 	&4.48 	&4.54	&3.28	&4.21    &3.17	&3.12	&3.08	&\textbf{2.84}	&\textbf{2.84}		\\
						&50\% 		&13.07	&9.95	&8.83	&9.08  &8.64	&8.74	&8.31	&8.16	&\textbf{7.98}	\\ \hline

\multirow{2}{*}{Satimage}&10\% 		&39.6 	&23.34	&14.38	&20.62  &17.32	&15.5	&14.7	&\textbf{13.06}	&14.24		\\
						&50\% 		&44.2	&24.24	&16.96	&24.1  &18.44	&16.18	&15.84	&\textbf{14.82}	&15.18	\\ \hline
\end{tabular}
\end{table*}

\subsection{Subspace clustering with missing entries}
\begin{figure}[h!]
\includegraphics[width=7cm]{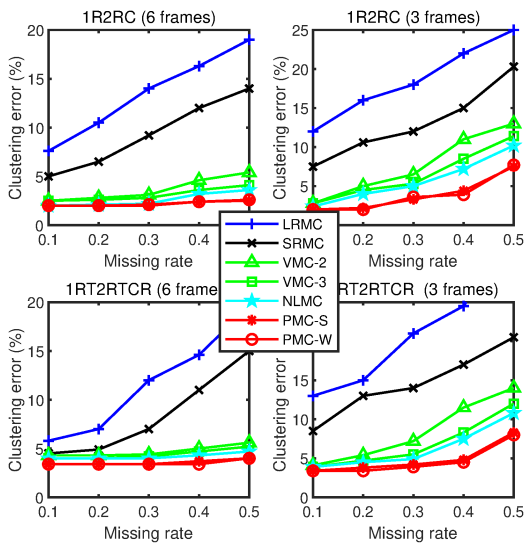}
\centering
\caption{Clustering errors of SSC on Hopkins 155 data recovered by different matrix completion methods}\label{fig.hopkins}
\end{figure}

Similar to \cite{YangRobinsonVidal,pmlr-v70-ongie17a}, we perform subspace clustering with missing data on the Hopkins 155 dataset \cite{4269999}. We consider two sequences of video frames, 1R2RC and 1RT2RTCR. For each sequence, we uniformly subsample $6$ and $3$ frames to form a high-rank matrix and a full-rank matrix respectively. We randomly remove some entries of the four matrices and perform matrix completion, followed by sparse subspace clustering \cite{SSC_PAMIN_2013}. The clustering errors (averages of 20 repeated trials) are shown in Figure \ref{fig.hopkins}, in which the missing rate denotes the proportion of missing entries. It can be found that PMC-S and and PMC-W consistently outperformed other methods in all cases.

\subsection{Motion capture data recovery}
\begin{figure}[h]
\includegraphics[width=6.5cm]{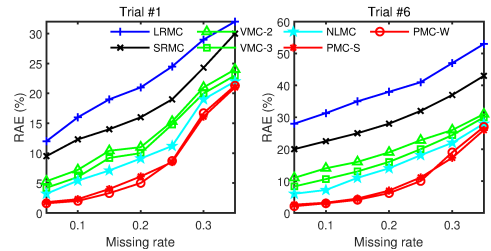}
\centering
\caption{RAE on CMU motion capture data}\label{fig.cmu}
\end{figure}

Similar to \cite{NIPS2016_6357,pmlr-v70-ongie17a}, we consider matrix completion on motion capture data, which consists of time-series trajectories of human motions such as running, jumping, and so on. We use the trials $\#1$ and $\#6$ of subject $\#56$ of the CMU Mocap dataset. We downsample the two datasets with factor 4. Then the sizes of the corresponding matrices are $62\times 377$ and $62\times 1696$. We randomly remove some fractions of the entries. Since the 62 features have different scales (the range of their standard deviations is about $(0,100)$), we use the relative absolute error (RAE) instead of RSE for evaluation:
$\textup{RAE}:=\sum_{(i,j)\in{\bar{\Omega}}}\vert\bm{X}_{ij}-\bm{\hat{X}}_{ij}\vert/\sum_{(i,j)\in\bar{\Omega}}\vert\bm{X}_{ij}\vert$.
The average results of 10 repeated trials are reported in Figure \ref{fig.cmu}. The recovery errors of our PMC-S and PMC-W are much lower than those of other methods.

\subsection{Transductive learning on incomplete data}
We consider four real datasets with missing values. More details and results are in the supplement. For each dataset, we randomly remove a fraction (denoted by $\theta$) of the observed entries of the feature matrix and perform classification. The proportion of training (labeled) data is $50\%$. We use matrix completion to recover the missing entries and classify the data simultaneously. The classification errors (average of 20 repeated trials) are reported in Table \ref{Tab_class_results}, in which the least classification error in each case is highlighted by bold. The classification error of SVM with zero-filled missing entries is very high if the missing rate is high. With the pre-processing of LRMC, the classification error of SVM can be significantly reduced. The classification errors of LRMC and SRMC are quite high because they are linear methods that are not effective in nonlinear classification. Our PMC-S and PMC-W outperform other methods in almost all cases.

\section{Conclusion}
In this paper we studied the problem of high-rank matrix completion and proposed a new model of matrix completion, which is based on analytic functions. We proposed two matrix completion methods PMC-S and PMC-W that outperformed state-of-the-art methods in subspace clustering with missing data, motion capture data recovery, and incomplete data classification. We analyzed the mechanism of the nonlinear recovery model and verified the superiority of rank minimization in kernel-induced feature space over rank minimization in data space theoretically and empirically. Future work may focus on improving the scalability of PMC.

\section{ Acknowledgments}
The authors gratefully acknowledge support from DARPA Award FA8750-17-2-0101 and NSF CCF-1740822.

\fontsize{9.5pt}{10.3pt}\selectfont

\bibliography{5948_Ref_PMC}
\bibliographystyle{aaai}

\end{document}